\documentclass{INTERSPEECH2023}

\interspeechcameraready

\title{Reducing Barriers to Self-Supervised Learning: \\HuBERT Pre-training with Academic Compute}
\name{William Chen$^1$, Xuankai Chang$^1$, Yifan Peng$^2$, Zhaoheng Ni$^3$, Soumi Maiti$^1$, Shinji Watanabe$^1$}
\address{
  $^1$Language Technologies Institute, Carnegie Mellon University, USA\\
  $^2$Electrical and Computer Engineering, Carnegie Mellon University, USA \\
  $^3$Meta, USA}
\email{\{wc4, xuankaic\}@andrew.cmu.edu}

\usepackage[
backend=biber,
style=ieee,
citestyle=numeric-comp,
maxbibnames=3,
maxcitenames=3,
doi=false,isbn=false,url=false,eprint=false
]{biblatex}
\usepackage{multicol}
\usepackage{multirow}

\defbibheading{bibliography}[\refname]{}

\DeclareSourcemap{
	\maps[datatype=bibtex, overwrite=true]{
		\map{
			\step[fieldsource=booktitle,
			match=\regexp{.*Interspeech.*},
			replace={Proc. Interspeech}]
			\step[fieldsource=journal,
			match=\regexp{.*INTERSPEECH.*},
			replace={Proc. Interspeech}]
			\step[fieldsource=booktitle,
			match=\regexp{.*ICASSP.*},
			replace={Proc. ICASSP}]
			\step[fieldsource=booktitle,
			match=\regexp{.*icassp_inpress.*},
			replace={Proc. ICASSP (in press)}]
			\step[fieldsource=booktitle,
			match=\regexp{.*Acoustics,.*Speech.*and.*Signal.*Processing.*},
			replace={Proc. ICASSP}]
			\step[fieldsource=booktitle,
			match=\regexp{.*International.*Conference.*on.*Learning.*Representations.*},
			replace={Proc. ICLR}]
			\step[fieldsource=booktitle,
			match=\regexp{.*International.*Conference.*on.*Computational.*Linguistics.*},
			replace={Proc. COLING}]
			\step[fieldsource=booktitle,
			match=\regexp{.*SIGdial.*Meeting.*on.*Discourse.*and.*Dialogue.*},
			replace={Proc. SIGDIAL}]
			\step[fieldsource=booktitle,
			match=\regexp{.*International.*Conference.*on.*Machine.*Learning.*},
			replace={Proc. ICML}]
			\step[fieldsource=booktitle,
			match=\regexp{.*North.*American.*Chapter.*of.*the.*Association.*for.*Computational.*Linguistics:.*Human.*Language.*Technologies.*},
			replace={Proc. NAACL}]
			\step[fieldsource=booktitle,
			match=\regexp{.*Empirical.*Methods.*in.*Natural.*Language.*Processing.*},
			replace={Proc. EMNLP}]
			\step[fieldsource=booktitle,
			match=\regexp{.*Association.*for.*Computational.*Linguistics.*},
			replace={Proc. ACL}]
			\step[fieldsource=booktitle,
			match=\regexp{.*Automatic.*Speech.*Recognition.*and.*Understanding.*},
			replace={Proc. ASRU}]
			\step[fieldsource=booktitle,
			match=\regexp{.*Spoken.*Language.*Technology.*},
			replace={Proc. SLT}]
			\step[fieldsource=booktitle,
			match=\regexp{.*Speech.*Synthesis.*Workshop.*},
			replace={Proc. SSW}]
			\step[fieldsource=booktitle,
			match=\regexp{.*workshop.*on.*speech.*synthesis.*},
			replace={Proc. SSW}]
			\step[fieldsource=booktitle,
			match=\regexp{.*Advances.*in.*neural.*information.*processing.*},
			replace={Proc. NeurIPS}]
			\step[fieldsource=booktitle,
			match=\regexp{.*Advances.*in.*Neural.*Information.*Processing.*},
			replace={Proc. NeurIPS}]
			\step[fieldsource=booktitle,
			match=\regexp{.*Workshop.*on.* Applications.* of.* Signal.*Processing.*to.*Audio.*and.*Acoustics.*},
			replace={Proc. WASPAA}]
			\step[fieldsource=publisher,
			match=\regexp{.+},
			replace={{}}]
			\step[fieldsource=month,
			match=\regexp{.+},
			replace={{}}]
			\step[fieldsource=location,
			match=\regexp{.+},
			replace={{}}]
			\step[fieldsource=address,
			match=\regexp{.+},
			replace={{}}]
			\step[fieldsource=organization,
			match=\regexp{.+},
			replace={{}}]
		}
	}
}
\begin{document}

\maketitle
 
\begin{abstract}
Self-supervised learning (SSL) has led to great strides in speech processing. However, the resources needed to train these models has become prohibitively large as they continue to scale. Currently, only a few groups with substantial resources are capable of creating SSL models, which harms reproducibility. In this work, we optimize HuBERT SSL to fit in academic constraints. We reproduce HuBERT independently from the original implementation, with no performance loss. Our code and training optimizations make SSL feasible with only 8 GPUs, instead of the 32 used in the original work. We also explore a semi-supervised route, using an ASR model to skip the first pre-training iteration. Within one iteration of pre-training, our models improve over HuBERT on several tasks. Furthermore, our HuBERT Large variant requires only 8 GPUs, achieving similar performance to the original trained on 128. As our contribution to the community, all models, configurations, and code are made open-source in ESPnet.

\end{abstract}

\section{Introduction}

Self-supervised pre-training has reached new epochs in recent years by scaling Transformer-based models \cite{vaswani2017attention}. By increasing both the model size and training data, impressive performance has been achieved across a wide variety of tasks \cite{srivastava2022beyond, wang-etal-2018-glue, wangSuperglue, yang21c_interspeech, imagenet, sslASRReview, sslReview}. In speech processing, self-supervision has been used at scale to learn powerful representations of acoustic features \cite{schneider19_wav2vec, baevskiw2v, hsuHubert, ChenWavLm, wangilsssl}. One promising approach to self-supervised learning (SSL) in speech is masked prediction-based training. Inspired by deep clustering \cite{caron2018deepCluster} and masked language modelling \cite{devlin-etal-2019-bert}, HuBERT \cite{hsuHubert} learns to predict the cluster assignment of masked regions of the acoustic input. Importantly, HuBERT requires multiple iterations of pre-training, using clusters derived from the features of previous iterations.

The computational expenses involved with pre-training current SSL models like HuBERT presents a major hurdle to reproducibility and the advancement of open science. Recent approaches are limited to using compression-based techniques such as distillation~\cite{distilhubert, fithubert, ashiharaDeepvsWide} and pruning~\cite{laiParp, pengStructuredPruning}, which require pre-trained SSL weights and cannot create a new standalone model. This makes SSL pre-training \textit{prohibitively expensive} to research groups without large-scale computing resources, like the 128 GPUs needed to train the original HuBERT Large. Unfortunately, the trend towards scaling up these models shows no sign of slowing down, which means that the ability to train them will become even more concentrated in the hands of a select few, further hampering the democratization of knowledge.

In the field of Natural Language Language Processing (NLP), where Large Language Models (LLMs) with billions of parameters \cite{chowdhery2022palm, brownGpt} are more common, attempts have been made to address this problem. For example, GPT-NeoX-20B \cite{black2022gptneoxb}, and BLOOM \cite{scao2022bloom} were the results of valuable community-driven efforts to build open-source LLMs. The benefits of this are manifold: researchers are able to share important optimization tricks and ablation logs, therefore reducing the amount of compute required for future studies. However, all of these works are limited to the NLP domain. To the best of our knowledge, such an effort has not been realized in speech.

Our goal is to build a pre-trained SSL model for speech processing without access to large-scale computing resources. Specifically, we want to reproduce HuBERT-style pre-training within a resource constraint that is feasible for academic settings. In doing so, we hope to share valuable insights for training SSL models with the broader research community, and lift the computational barriers that have recently surrounded large-scale self-supervised training.

We approach this challenge from two main directions. The first is to reproduce HuBERT pre-training such that SSL is feasible on a limited compute budget. This includes a multitude of factors, such as storage space, pre-processing time, training time, and most critically, the number of GPUs required for training. The other angle is to reduce the compute necessary to train a standalone SSL model. We explore a semi-supervised SSL (semi-SSL) approach, where the first HuBERT iteration is replaced with a supervised model trained on Automatic Speech Recognition (ASR). In doing so, the costs for SSL pre-training can be drastically lowered, even when considering the costs of training the supervised model.

Overall, we showed that the computational overhead of SSL pre-training can be considerably reduced. Due to our optimization of the code, pre-processing, and training settings, we are able to achieve shorter pre-training times with only 8 GPUs than that of prior work \cite{yang2022torchaudio, hsuHubert}, which used 32. Our reproduced HuBERT achieves comparable results to the original implementation across the SUPERB benchmark \cite{yang21c_interspeech}. Furthermore, our semi-SSL models, pre-trained with only 8 GPUs and for a single iteration, improve over comparable state-of-the-art SSL models on 4 different SUPERB tasks. We open source all of our models, training configurations, and code as one of our core contributions, so that different implementations are available to the research community for SSL.

\vspace{-0.2cm}
\section{Model Design and Implementation}

In this section, we first provide background on HuBERT-based SSL and discuss our reproduction efforts. We then outline our semi-SSL approach, designed to reduce the amount of HuBERT training iterations. Afterwards, we discuss modifications made to HuBERT pre-processing and training, allowing for it to be done even in more resource constrained settings.
\vspace{-0.2cm}
\subsection{Reproducing HuBERT} \label{sec:reproduce}

HuBERT builds off of the wav2vec 2.0 architecture \cite{baevskiw2v}, consisting of a convolutional feature extractor, a Transformer encoder, and a codebook output layer.  The labels used to train the model are a sequence of cluster assignments. The assignments are obtained using k-means clustering on the acoustic features, where $k$ is equal to the codebook size. The acoustic features are Mel-Frequency Cepstral Coefficients (MFCCs) or the encoded representations of a pre-trained model. HuBERT is then trained to predict the cluster assignments of masked regions in the audio. We reproduce HuBERT following the original pipeline \cite{hsuHubert}. Iteration 0 is pre-trained on clusters derived from 39-dimension MFCC features extracted from LibriSpeech 960h \cite{librispeech}, which are clustered with a k-means codebook size of 100 for pseudo-labeling. Iteration 1 is then trained on features extracted from the 6th layer of the iteration 0 model clustered with a codebook size of 500. Both iterations use the HuBERT Base architecture. Finally, iteration 2 uses the HuBERT Large architecture and is trained on clusters derived from the 9th layer of the iteration 1 model, using a cluster size of 500.

One issue we found when reproducing HuBERT was that there was little correlation between the pre-training validation curves and fine-tuning performance. For example, a model checkpoint after 400K steps had wildly different ASR results from a checkpoint at 800K steps, despite very similar validation loss/accuracy. This meant that hyperparameter tuning required full pre-training trials, each of which required close to \~1000 GPU hours. We conducted 10 trials in total, spending a total of 9600 GPU hours, before finding an optimal setting that matched the original in performance. These settings are shared with the community, so that future efforts can be accelerated.

\subsection{Semi-Self-Supervised Learning}
\label{sec:hubert_feats}

\begin{figure}[tb]
    \centering
    \includegraphics[height=2.5cm]{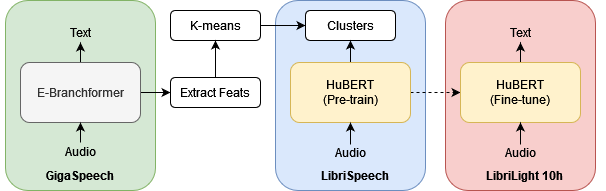}
    \caption{Diagram of our model pipeline. The supervised E-Branchformer is used to extract hidden representations, which are then clustered with k-means. The HuBERT model is pre-trained on these clusters, before being fine-tuned.}
    \label{fig:diagram}
    \vskip -0.4cm
\end{figure}
\label{sec: method}

As mentioned above, SSL training requires significant trial and error due to the lack of direct evaluation metrics. This is most critical for iteration 0 of HuBERT, which serves as the foundation for subsequent iterations. On the other hand, supervised ASR models are much less costly to train and tune, as they can be directly evaluated through fine-tuning metrics, such as validation loss and accuracy \footnote{While these may not be perfectly indicative of final performance, they are still more insightful than HuBERT SSL metrics, such as masked prediction accuracy, which depend on the quality/number of the target clusters. Trivial clusters may lead to high masked prediction accuracy but poor downstream results.}. Could we instead replace the entire first iteration of HuBERT pre-training with supervised training? In doing so, we could eliminate the lengthy hyper-parameter tuning process of an entire pre-training iteration.

We explore a semi-supervised SSL (semi-SSL) route, where a supervised model is used replace iteration 0 of pre-training (Figure \ref{fig:diagram}). Instead of using MFCC features for the initial clusters, we extract the hidden representations of an ASR model. Since the goal is to reduce the amount of training iterations as much as possible, our approach is to use the strongest possible ASR model that can yield better initial clusters. Our choice is a powerful ASR encoder-decoder, built on top of the state-of-the-art E-Branchformer \cite{kimEbranch} architecture, and trained on the large-scale GigaSpeech corpus \cite{gigaspeech}. One important decision was determining which layer of the ASR model to extract the features from.  We selected features from the middle layer (layer 8) for one pre-training trial, since they had the highest phonetic content when evaluated against frame-level phonetic transcripts \cite{mcauliffe17Aligner, LugoschLibrispeehAlign}. We conducted an additional trial using features from the last layer (layer 16), which was found to encode the most linguistic content \cite{shim2022understanding}.

While our methodology is similar to recent developments in task-oriented pre-training, the core motivations are different. Prior works \cite{wangPbert, asbert, kreyssig2022biased} adopted a self-training approach, where supervised models are boosted via unsupervised pre-training. A fine-tuned ASR model is used to initialize SSL, and then fine-tuned again on the same supervised data. Their goal is to improve performance on a \textit{specific} task and domain. In doing so, the performance of the model may be degraded when used for other tasks, such as with the case of speaker verification \cite{wangPbert}. However, our motivation is to reduce the cost of pre-training while maintaining performance across a \textit{range} of tasks. Because of this, it is important that our teacher ASR model is trained on a broad range of data, so that the semi-SSL model can be usable across a wide variety of tasks and domains.

\subsection{Optimizing Pre-processing} 
Most costs associated with SSL occur during the pre-training stage, but we also found that pre-processing also required a significant investment. The official implementation of HuBERT in fairseq \cite{ott2019fairseq} saves the extracted features necessary for k-means to the disk as NumPy \cite{numpy} arrays. All of the features were loaded into a single node for k-means training. This led to large amounts of memory usage, particularly when using the high dimensional hidden representations. To alleviate this, we implemented a k-means algorithm parallelized across multiple computing nodes. We further optimized the k-means process by swapping NumPy for Kaldiio \footnote{https://github.com/nttcslab-sp/kaldiio}. This allows the extracted features to be sampled from k-means training prior to loading from disk (rather than after as in the original implementation), reducing the amount of wasted memory and I/O time. On ~96 hours of extracted 39-dimension MFCC features, we are able to reduce the maximum memory usage during k-means from 27.3GB to 162MB. The reductions for k-means on the 512-dimension ASR hidden representations are even more significant: memory usage is reduced from 122GB to 4.6 GB. 

One major constraint was storage space. The audio used to train the HuBERT-Large model consumed over 7TB of disk space, with the majority from Libri-Light 60k \cite{kahnLibriLight}. The feature extraction step in HuBERT only exacerbated the problem due to their high dimensionality. We found that extracting all of the features for Libri-Light would consume 25TB of storage. To avoid this, we made significant modifications to the clustering pipeline. The k-means model was trained using features from LibriSpeech, which only used 500GB of space. The Libri-Light features would then be converted into cluster assignments with the trained k-means model within memory, removing the need to store the extracted features. Through this, we reduced the storage footprint of these steps from 25TB to 500GB. 

\subsection{Training Optimizations}
\vspace{-0.2cm}
During training, we only had access to 8 GPUs, one-fourth of the 32 GPUs used in the original implementation. However, the authors of HuBERT also found batch size to be a large determinant in model performance \cite{hsuHubert}. To make up for the contrast in resources, we use gradient accumulation to increase our effective batch size to similar levels. Furthermore, we utilized mixed precision training with brain float 16 to increase training speed. 
\vspace{-0.4cm}
\section{Experimental Setup}

For independent reproducibility, we conduct all experiments outside of the original fairseq \cite{ott2019fairseq} framework, implementing our full pipeline within ESPnet \cite{watanabe2018espnet}.
\vspace{-0.2cm}
\subsection{Datasets}
\vspace{-0.2cm}
We use GigaSpeech \cite{gigaspeech}, a 10k hour corpus of English speech across a variety of domains, such as audiobooks and podcasts, to train the ASR model. We use the 960 hours of LibriSpeech \cite{librispeech} audio to pre-train all models using the HuBERT Base architecture. Our models that used the HuBERT Large architecture is trained on 60k hours of audio from Libri-Light \cite{kahnLibriLight}. For model fine-tuning, we used the 10 hours labeled subset of Libri-Light. To understand the generalizability of our semi-SSL models, we also evaluate them using the SUPERB Benchmark \cite{yang21c_interspeech}.
\vspace{-0.2cm}
\subsection{HuBERT} \label{sec: hubert}
\vspace{-0.2cm}
We reproduce HuBERT following the original pipeline \cite{hsuHubert} as described in Section \ref{sec:reproduce}. Iterations 0 and 1 follow the HuBERT Base architecture, which consists of 12 encoder layers with a hidden size of 768.  The models were trained using the Adam optimizer for 880k steps. We used a warmup learning rate scheduler, such that the learning rate was linearly increased for 32k steps until it reached a peak of 0.0005, before exponentially decaying. Mini-batches are constructed using ESPnet's \texttt{numel} sampler, using a batch bins size of 45M to maximize the data in each GPU. Hyper-parameter search was conducted throughout 10 full trials of pre-training. Iteration 2 follows the HuBERT Large architecture, which has 24 encoder layers and a hidden size of 1024. The model was trained with the same optimization settings as the Base-sized models. Due to the size of the model, we reduce the batch size to one audio clip per GPU.

\vspace{-0.2cm}
\subsection{Semi-supervised SSL Pre-training}
\vspace{-0.2cm}
\subsubsection{Supervised ASR Teacher}
\vspace{-0.2cm}
The ASR model is a hybrid CTC/attention \cite{watanabeHybrid} encoder-decoder. The encoder is a 17-layer E-Branchformer \cite{kimEbranch} (an enhanced version of Branchformer~\cite{pengBranch}), while the decoder is a 6-layer Transformer \cite{vaswani2017attention}. We use the same configuration as the existing E-Branchformer recipe for LibriSpeech. Each encoder layer has a hidden size of 512, 8 attention heads, and a kernel size of 31. Each decoder layer has 8 attention heads and a feed-forward dimension of 2048. 
\vspace{-0.2cm}
\subsubsection{Semi-SSL Models}
\vspace{-0.2cm}
As mentioned in Section \ref{sec:hubert_feats}, we first use the ASR teacher to extract representations of audio from LibriSpeech, for both layers 8 and 16. Clusters are then created using the k-means algorithm with a codebook size of 500. Due to the larger memory footprint from the increased output size, the batch bins size is reduced to 40M. All other settings remain the same as the reproduced HuBERT. Our models are denoted as the following:

\noindent \textbf{GS-8}: Iteration 0 clusters were derived from the middle layer (layer 8) of the ASR model, which contained the most phonetic content. We only train this model for iteration 0, using the HuBERT Base architecture.

\noindent \textbf{GS-16 Base}: Iteration 0 training targets were derived from the last layer (layer 16) of the ASR teacher, which should contain the most linguistic content \cite{shim2022understanding}. This model was trained for two iterations, both using the HuBERT Base architecture.

\noindent \textbf{GS-16 Large}: This model was trained on clusters obtained from iteration 0 of GS-16 Base. It uses the same settings as the HuBERT Large described in Section \ref{sec: hubert}
\vspace{-0.2cm}
\subsection{Fine-tuning}
\vspace{-0.2cm}
We follow the fine-tuning setup of \cite{yang2022torchaudio}\footnote{\label{torchaudionote}https://github.com/pytorch/audio/tree/main/examples/hubert}, allowing for comparisons in both compute usage and downstream performance. Models are fine-tuned on 10 hours of Libri-Light using CTC loss \cite{graves2006connectionist} for 10200 updates, with the pre-trained encoder frozen for the first 4000. We use a character level vocabulary of size 31. The 10 best checkpoints are averaged for inference, which is done with CTC greedy decoding and without a language model. 

\vspace{-0.2cm}
\subsection{Hardware}
\vspace{-0.2cm}
We train the majority of models on publicly accessible HPC clusters with 8xA100 GPUs. To conduct more experimental trials, we also train models using Amazon AWS EC2; we used a single 40GB A100x8 instance and 32GB V100x8 instance. 

\vspace{-0.2cm}
\section{Results} 
\begin{table}[tb]
    \centering
        \caption[]{Word Error Rate on Librispeech dev/test sets. SSL models were fine-tuned on 10 hours of data from Librilight, using CTC loss without a language model. Results from \cite{yang2022torchaudio} were obtained from the official repository \footnotemark{}.}
    
\begin{tabular}{lcccc}
\toprule
model  & \multicolumn{2}{c}{dev} & \multicolumn{2}{c}{test}\\ 
  & clean & other & clean & other \\ \midrule
 \multicolumn{5}{l}{\textit{Prior Work (Iteration 1)}} \\
HuBERT Base\cite{yang2022torchaudio} & 10.9 & 17.5 & 10.9 & 17.8 \\
\midrule
\multicolumn{5}{l}{\textit{Iteration 0}} \\
HuBERT Base & 15.5 & 23.9 & 15.9 & 24.1\\
GS-8 & 12.2 & 19.4 & 12.3 & 19.9 \\
GS-16 & \textbf{9.6} & \textbf{15.7} & \textbf{9.7} & \textbf{16.3} \\
\midrule
\multicolumn{5}{l}{\textit{Iteration 1}} \\
HuBERT Base & 10.4 & 17.4 & 10.5 & 17.6\\
GS-16 (Base) & 10.2 & 16.0 & 10.3 & 16.4 \\ 
GS-16 (Large) & \textbf{8.4} & \textbf{13.8} & \textbf{8.2} & \textbf{13.8} \\ 
\midrule
\multicolumn{5}{l}{\textit{Iteration 2}} \\
HuBERT Large & 9.5 & 14.9 & 9.6 & 14.9\\
\bottomrule
\end{tabular}

    \label{tab:librilight}
    \vskip -0.35in
\end{table}
\begin{table*}[htb]
    \centering
    \caption{Evaluation on the SUPERB Benchmark. The tasks are grouped into 5 categories: recognition, detection, semantics, speaker, and paralinguistics. Metrics for each task are: phoneme error rate (PR), word error rate (ASR), maximum term weighted value (QbE), F1 and concept error rate (SF), equal error rate (ASV), diarization error rate (SD), and accuracy (KS, IC, and ER)}
         \resizebox {\linewidth} {!} {

 \begin{tabular}{lcccccccccccc}
 \toprule
 \multirow{3}*[-2pt]{Method} & \multirow{3}*[-2pt]{Params} & \multirow{3}*[-2pt]{\begin{tabular}{@{}c@{}}Score\end{tabular}} & \multicolumn{2}{c}{Recognition} & \multicolumn{2}{c}{Detection}  & \multicolumn{3}{c}{Semantics} & \multicolumn{2}{c}{Speaker} & Paralinguistics \\
 \cmidrule(r){4-5}\cmidrule(r){6-7}\cmidrule(r){8-10}\cmidrule(r){11-12}\cmidrule(r){13-13}
 & & & PR$\downarrow$ & ASR$\downarrow$ & KS $\uparrow$ & QbE $\uparrow$  & IC $\uparrow$ & SF (F1) $\uparrow$ & SF (CER) $\downarrow$ & ASV $\downarrow$  & SD $\downarrow$ & ER $\uparrow$ \\
 \midrule
  \multicolumn{11}{l}{\textit{Prior work}} \\
 DistillHuBERT \cite{distilhubert} & 23.5M & 76.2 & 16.3 & 13.4 & 95.9 & 5.1 & 95.0 & 82.6 & 36.6 & 8.6 & 6.2 & 63.0    \\
 LightHuBERT \cite{wang22tLight} & 95.0M & 81.2 & 4.2 & 5.7 & \textbf{96.8} & 7.4 & 98.5 & 88.4 & 25.3  & 5.1 & \textbf{5.5} & 66.3 \\
 HuBERT Base \cite{hsuHubert} & 94.7M & 80.7 & 5.4 & 6.4 & 96.3 & 7.4 & 98.3 & 88.5 & 25.2  & 5.1 & 5.9 & 64.9 \\
 HuBERT Large \cite{hsuHubert} & 316M & 81.4 & \textbf{3.5} & \textbf{3.6} & 95.3 & \textbf{3.5} & \textbf{98.8} & \textbf{89.8} & \textbf{21.8} & 6.0 & 5.8 & \textbf{67.6} \\
 \midrule
\multicolumn{11}{l}{\textit{Iteration 0}} \\
GS-8 & 94.7M & 80.9 & 6.3 & 6.9 & \textbf{96.1} & \textbf{8.1}  & 98.9 & 89.4 & 22.5  & \textbf{5.9} & \textbf{6.2} & \textbf{64.1}\\
GS-16 & 94.7M & 80.8 & \textbf{6.1} & \textbf{5.4} & 95.6 &  5.6 & \textbf{99.1} & \textbf{90.0} & \textbf{21.5} & 6.4 & 6.5  & 64.0\\
\midrule
\multicolumn{11}{l}{\textit{Iteration 1}} \\
HuBERT Base & 94.7M & 80.7 & 5.8 & 6.9 & 95.8 & 10.1 & 97.6 & 89.0 & 24.1   & 5.6 & 6.1 &  62.8\\ 
GS-16 (Base) & 94.7M & 81.3 & \textbf{5.3} & 6.6 & 96.3 & \textbf{10.5} & \textbf{99.0} & \textbf{89.9} & \textbf{22.1}  & 5.6 & 6.2 & 63.1  \\
GS-16 (Large) & 316M & 81.4 & 7.4 & \textbf{5.4} & \textbf{97.4} & 10.1 & 98.9  & 89.0 & 23.3 & 6.4 & \textbf{5.6} & \textbf{66.1}\\
\midrule
\bottomrule
 \end{tabular}
 }
    \label{tab:superb}
    \vskip -0.3in
\end{table*}

\subsection{ASR Fine-tuning}
\vspace{-0.2cm}
The results are presented in Table \ref{tab:librilight}, evaluated using Word Error Rate (WER) on the LibriSpeech test sets. Our reproduced HuBERT Base yielded strong results by outperforming prior work \cite{yang2022torchaudio}, demonstrating the feasibility of SSL even with limited compute. Both GS-8 and GS-16 performed better than HuBERT Base in iteration 0, showing the effectiveness of semi-supervised SSL. We observed that GS-16 performed much better than GS-8, likely as a result of its clusters encoding content more relevant to ASR. Furthermore, GS-16 had a lower WER than our reproduced HuBERT Base, which was pre-trained for an additional iteration (0.8/1.3 WER lower on test-clean/other). \footnotetext{Ibid.}

\vspace{-0.5cm}
\subsection{Universal Representations Benchmark}
\vspace{-0.2cm}
SUPERB consists of several speech processing tasks designed to evaluate SSL models. Tasks include Phoneme Recognition (PR), ASR, Keyword Spotting (KS), Query by Example (QbE), Intent Classification (IC), Slot Filling (SF), Automatic Speaker Verification (ASV), Speaker Diarization (SD), and Emotion Recognition (ER). To obtain an overall picture of each model's performance, we calculate a score in the same way as \cite{ChenWavLm, wang22tLight}: QbE is multiplied by 100, error rates are subtracted from 1, and average the scores across all tasks.

The results on SUPERB are contained in Table \ref{tab:superb}. To compare with prior work, we include results for the original HuBERT implementation. We also include the scores of distillation-based approaches \cite{distilhubert, wang22tLight}, although they cannot be used to train a standalone SSL model. Our reproduced HuBERT performs at a similar level to the original, showing that our training optimizations did not degrade the performance. Our semi-SSL models also achieved comparable scores to the original HuBERT within a single iteration of pre-training (80.9 and 80.8 vs 80.7). These results indicate that features used for the initial cluster assignments had a significant effect on downstream performance. GS-16 had significant gains in semantic-related tasks and ASR. However, this came at the cost of performance in other categories, such as detection. On the other hand, GS-8 obtained scores that were more even across the board, having less degradation on detection-related tasks. 

The GS-16 models trained for an additional iteration achieved even stronger results. The overall score improved from 80.9 to 81.3 for the GS-16 Base model, which is close to the original HuBERT Large (81.4). Our GS-16 Large also matches the performance of the original HuBERT Large, while being trained for 1 less iteration. This shows that our semi-SSL technique is effective in reducing the computational expenses of SSL while maintaining performance.

\vspace{-0.2cm}
\subsection{Training Time}
\begin{table}[tb]
    \centering
    \caption{Models by compute usage. Our models consume less GPU resources while maintaining similar performance.}
    \vspace{-0.2cm}
      \resizebox {\linewidth} {!} {
\begin{tabular}{lcccc}
\toprule
Model  & GPU & Steps & GPU Hours & Days\\ \midrule
\multicolumn{5}{l}{\textit{Iteration 0}} \\
HuBERT Base \cite{yang2022torchaudio} & A100x32 & 250K & 1344 & 2\\
HuBERT Base (ours) & A100x8  & 880K & 960 & 5\\
GS-8 & A100x8  & 880K & 1166 & 6\\
GS-16 & A100x8   & 880K & 1166 & 6\\
\midrule
\multicolumn{5}{l}{\textit{Iteration 1}} \\
HuBERT Base \cite{yang2022torchaudio} & A100x32 & 400k & 2150 & 3\\
HuBERT Base (ours) & A100x8 & 880k & 960   & 5\\
GS-16 Base & V100x8 & 880k  & 1600  & 8\\  
GS-16 Large & A100x8 & 3M & 2880  & 15\\  
\midrule
\multicolumn{5}{l}{\textit{Iteration 2}} \\
HuBERT Large & A100x16 & 2.4M & 7550 & 20\\
\midrule
\bottomrule
\end{tabular}
}

    \label{tab:hardware}
    \vskip -0.3in
\end{table}
Table \ref{tab:hardware} breaks down the resources and time required to train each model. Each HuBERT Base iteration could be trained  in 960 to 1200 GPU hours using 8 Nvidia A100s. We highlight this significant improvement over prior work, which consumed 1344 to 2150 GPU hours \cite{yang2022torchaudio}. Specifically, they required 16.8 hours for every 100K gradient steps when training a HuBERT Base model on 32 A100 GPUs. Our optimizations resulted in a significant speedup, such that the same amount of steps only required \~16.1 hours with 8 GPUs.

\vspace{-0.3cm}
\section{Conclusion}
\vspace{-0.1cm}
\label{sec: conclusion}
Advances in SSL have focused on scaling larger models to more data. While effective, this constrains SSL research to only those that have considerable computing resources. Our work focuses on training state-of-the-art SSL models within academic resource constraints. We reproduced HuBERT independently from the original implementation, while optimizing its training to fit within only 8 GPUs, and without performance loss. To further reduce costs, we leverage supervised models to initialize the clusters assignments for a semi-SSL HuBERT. Our evaluation on SUPERB shows that semi-SSL models can reach comparable, if not better, results across many tasks within a single iteration of pre-training. Furthermore, we showcase the feasibility of academic SSL by pre-training a 300M parameter semi-SSL model using only 8 GPUs, as compared to the original 128, which yields results comparable to HuBERT Large. All of our models, training configurations, and code are made available open-source as our core contribution to the community.

\vspace{-0.2cm}
\section{Acknowledgements}
This work used the PSC Bridges2 and the NCSA Delta systems from the Advanced Cyberinfrastructure Coordination Ecosystem, as part of project cis210027p and supported by NSF award number ACI-1445606.

\clearpage
\section{References}
{
\printbibliography
}
\end{document}